\def\year{2020}\relax
\documentclass[letterpaper]{article} 
\usepackage{aaai20}  
\usepackage{times}  
\usepackage{helvet} 
\usepackage{courier}  
\usepackage[hyphens]{url}  
\usepackage{graphicx} 
\usepackage{graphicx}  
\usepackage[utf8]{inputenc} 
\usepackage[T1]{fontenc}    
\usepackage{url}            
\usepackage{booktabs}       
\usepackage{amsfonts}       
\usepackage{nicefrac}       
\usepackage{enumitem}
\usepackage{multirow}
\usepackage{array}

\usepackage{times}
\usepackage{soul}
\usepackage{graphicx}
\usepackage{subfigure}
\usepackage{amsmath,amsthm}
\usepackage{algorithm}
\usepackage{algorithmic}
\usepackage{amssymb}
\theoremstyle{plain}
\newtheorem{theorem}{Theorem}
\newtheorem{corollary}{Corollary}

\newtheorem{lemma}{Lemma}
\newtheorem{question}{Q}

\theoremstyle{definition}

\newtheorem{model}{Model}

\DeclareMathOperator*{\concat}{\scalebox{1}[1.0]{$\parallel$}}

\title{Improving Attention Mechanism in Graph Neural Networks via Cardinality Preservation}

\author{
Shuo Zhang,\textsuperscript{1} Lei Xie\textsuperscript{1,2,3} \\
\textsuperscript{1}Ph.D. Program in Computer Science, The Graduate Center, The City University of New York \\
\textsuperscript{2}Department of Computer Science, Hunter College, The City University of New York \\
\textsuperscript{3}Helen and Robert Appel Alzheimer’s Disease Research Institute, Feil Family Brain and Mind Research Institute, \\Weill Cornell Medicine, Cornell University \\
szhang4@gradcenter.cuny.edu, lei.xie@hunter.cuny.edu \\
}

\begin{document}

\maketitle

\begin{abstract}
Graph Neural Networks (GNNs) are powerful to learn the representation of graph-structured data. Most of the GNNs use the message-passing scheme, where the embedding of a node is iteratively updated by aggregating the information of its neighbors. To achieve a better expressive capability of node influences, attention mechanism has grown to be popular to assign trainable weights to the nodes in aggregation. Though the attention-based GNNs have achieved remarkable results in various tasks, a clear understanding of their discriminative capacities is missing. In this work, we present a theoretical analysis of the representational properties of the GNN that adopts the attention mechanism as an aggregator. Our analysis determines all cases when those attention-based GNNs can always fail to distinguish certain distinct structures. Those cases appear due to the ignorance of cardinality information in attention-based aggregation. To improve the performance of attention-based GNNs, we propose cardinality preserved attention (CPA) models that can be applied to any kind of attention mechanisms. Our experiments on node and graph classification confirm our theoretical analysis and show the competitive performance of our CPA models.

\end{abstract}

\section{Introduction}
Graph, as a kind of powerful data structure in non-Euclidean domain, can represent a set of instances (nodes) and the relationships (edges) between them, thus has a broad application in various fields~\cite{zhou2018graph}. Different from regular Euclidean data such as texts, images and videos, which have clear grid structures that are relatively easy to generalize fundamental mathematical operations~\cite{shuman2013emerging}, graph structured data are irregular so it is not straightforward to apply important operations in deep learning (e.g. convolutions). Consequently, the analysis of graph-structured data remains a challenging and ubiquitous question.

In recent years, Graph Neural Networks (GNNs) have been proposed to learn the representations of graph-structured data and attract a growing interest~\cite{scarselli2009graph,li2015gated,duvenaud2015convolutional,niepert2016learning,kipf2017semi,hamilton2017inductive,zhang2018end,ying2018hierarchical,morris2018weisfeiler,xu2018how}. GNNs can iteratively update node embeddings by aggregating/passing node features and structural information in the graph. The generated node embeddings can be fed into an extra classification/prediction layer and the whole model is trained end-to-end for different tasks.

Though many GNNs have been proposed, it is noted that when updating the embedding of a node ${v_i}$ by aggregating the embeddings of its neighbor nodes ${v_j}$, most of the GNN variants will assign non-parametric weight between ${v_i}$ and ${v_j}$ in their aggregators~\cite{kipf2017semi,hamilton2017inductive,xu2018how}. However, such aggregators (e.g. sum or mean) fail to learn and distinguish the information between a target node and its neighbors during the training. Taking account of different contributions from the nodes in a graph is important in real-world data as not all edges have similar impacts. A natural alternative solution is making the edge weights trainable to have a better expressive capability.

To assign learnable weights in the aggregation, attention mechanism~\cite{bahdanau2014neural,vaswani2017attention} is incorporated in GNNs. Thus the weights can be directly represented by attention coefficients between nodes and give interpretability~\cite{velickovic2018graph,thekumparampil2018attention,zhou2018commonsense}. Though GNNs with the attention-based aggregators achieve promising performance on various tasks empirically, a clear understanding of their discriminative power is missing for the designing of more powerful attention-based GNNs. Recent works~\cite{morris2019weisfeiler,xu2018how,maron2019provably} have theoretically analyzed the expressive power of GNNs. However, they are unaware of the attention mechanism in their analysis. So that it's unclear whether using attention mechanism in aggregation will constrain the expressive power of GNNs.

In this work, we make efforts to theoretically analyze the discriminative power of GNNs with attention-based aggregators. Our findings reveal that previous proposed attention-based aggregators fail to distinguish certain distinct structures. By determining all such cases, we reveal the reason for those failures is the ignorance of cardinality information in aggregation. It inspires us to improve the attention mechanism via cardinality preservation. We propose models that can be applied to any kind of attention mechanisms and achieve the goal. In our experiments on node and graph classifications, we confirm our theoretical analysis and validate the power of our proposed models. The best-performed one can achieve competitive results comparing to other baselines. Specifically, our key contributions are summarized as follows:
\begin{itemize}
\item We show that previously proposed attention-based aggregators in message-passing GNNs always fail to distinguish certain distinct structures. We determine all of those cases and demonstrate the reason is the ignorance of the cardinality information in attention-based aggregation.
\item We propose Cardinality Preserved Attention (CPA) methods to improve the original attention-based aggregator. With them, we can distinguish all cases that previously always fail an attention-based aggregator.
\item Experiments on node and graph classification validate our theoretical analysis and the power of our CPA models. Comparing to baselines, CPA models can reach state-of-the-art level.
\end{itemize}

\section{Preliminaries}

\subsection{Notations}
Let $G = ( V , E )$ be a graph with set of nodes $V$ and set of edges $E$. The nearest neighbors of node $i$ are defined as $\mathcal { N } ( i ) = \{ j | d ( i , j ) = 1 \}$, where $d ( i , j )$ is the shortest distance between node $i$ and $j$. We denote the set of node $i$ and its nearest neighbors as $\tilde{\mathcal { N }}( i ) = \mathcal { N } ( i ) \cup \{ i \}$. For the nodes in $\tilde{\mathcal { N }} ( i )$, their feature vectors form a {\em multiset} $M(i) = (S_i, \mu_i )$, where $S_i=\left\{ s_1, \ldots, s_n \right\}$ is the {\em ground set} of $M(i)$, and $\mu_i : S_i \rightarrow \mathbb{N}^{*}$ is the {\em multiplicity function} that gives the {\em multiplicity} of each $s \in S_i$. The {\em cardinality} $| M |$ of a multiset is the number of elements (with multiplicity) in the multiset.

\subsection{Graph Neural Networks}

\subsubsection{General GNNs} Graph Neural Networks (GNNs) adopt element (node or edge) features $X$ and the graph structure $A$ as input to learn the representation of each element, $h _ { i }$, or each graph, $h _ { G }$, for different tasks. In this work, we focus on the GNNs under massage-passing framework, which updates the node embeddings by aggregating its nearest neighbor node embeddings iteratively. In previous surveys, this type of GNNs is referred as Graph Convolutional Networks in~\cite{wu2019comprehensive} or the GNNs with convolutional aggregator in~\cite{zhou2018graph}. Under the framework, a learned representation of the node after $l$ aggregation layers can contain the features and the structural information within $l$-step neighborhoods of the node. The $l$-th layer of a GNN can be formally represented as:
\begin{align}
h _ { i } ^ { l } = \phi ^ { \, l } \big( h _ { i } ^ { l - 1 } , \left\{ h _ { j } ^ { l - 1 }, \forall  j \in \mathcal { N } ( i ) \right\} \big), \label{aggregate}
\end{align}
where the superscript $l$ denotes the $l$-th layer and $h _ { i } ^ { 0 }$ is initialized as $X _ { i } $. The aggregation function $\phi$ in equation \ref{aggregate} propagates information between nodes and updates the hidden state of nodes.

In the final layer, since the node representation $h _ { i } ^ { L }$ after $L$ iterations contains the $L$-step neighborhood information, it can be directly used for local/node-level tasks. While for global/graph-level tasks, the whole graph representation $h _ { G }$ is needed, which requiring an extra readout function $g$ to compute $h _ { G }$ from all $h _ { i } ^ { L }$:
\begin{align}
h _ { G } = g \big( \left\{ h _ { i } ^ { L }, \forall i \in G \right\} \big).\label{READOUT}
\end{align}

\subsubsection{Attention-Based GNNs} In a GNN, when the aggregation function $\phi$ in equation \ref{aggregate} adopts attention mechanism, we consider it as an attention-based GNN. In previous survey (Section 6 of~\cite{lee2018attention}), this is referred to the first two types of attentions which have been applied to graph data. The attention-based aggregator in $l$-th layer can be formulated as follows:
\begin{align}
& e _ { i j }^ { l-1 } = Att \big( h _ { i }^ { l-1 } , h _ { j }^ { l-1 } \big),\label{attention}\\
& \alpha _ { i j }^ { l-1 } = \operatorname { softmax } \big( e _ { i j }^ { l-1 } \big) = \frac { \exp ( e _ { i j }^ { l-1 } ) } { \sum _ { k \in \tilde{\mathcal { N }}( i ) } \exp \big( e _ { i k }^ { l-1 } \big) },\label{softmax}\\
& h _ { i } ^ { l } =  f ^ { l } \Big( \sum\nolimits _ { j \in \tilde{\mathcal { N }} ( i )}\alpha _ { i j } ^ { l-1 } h _ { j } ^ {  l-1 } \Big), \label{weight-sum}
\end{align}
where the superscript $l$ denotes the $l$-th layer and $e _ { i j }$ is the attention coefficient computed by an attention function $Att$ to measure the relation between node $i$ and node $j$. $\alpha _ { i j }$ is the attention weight calculated by the softmax function. Equation \ref{weight-sum} is a weighted summation that uses all $\alpha$ as weights followed with a nonlinear function $f$.

\subsection{Related Works}
Since GNNs have achieved remarkable results in practice, a clear understanding of the power of GNNs in graph representational learning is needed to design better models and make further improvements. Recent works~\cite{morris2019weisfeiler,xu2018how,maron2019provably} focus on understanding the discriminative power of GNNs by comparing it to the Weisfeiler-Lehman (WL) test~\cite{weisfeiler1968reduction} when deciding the graph isomorphism. It is proved that massage-passing-based GNNs which aggregate the nearest neighbor node features of a node for embedding are at most as powerful as the 1-WL test~\cite{xu2018how}. Inspired by the higher discriminative power of the $k$-WL test ($k > 2$)~\cite{cai1992optimal} than the 1-WL test, GNNs that have a theoretically higher discriminative power than the massage-passing-based GNNs have been proposed based on the $k$-WL test~\cite{morris2019weisfeiler,maron2019provably}. However, the GNNs proposed in those works don't specifically contain the attention mechanism as the part of their analysis. So it's currently unknown whether the attention mechanism will constrain the discriminative power. Our work focuses on the massage-passing-based GNNs with attention mechanism, which are upper bounded by the 1-WL test.

Another recent work~\cite{knyazev2019understanding} aims to understand the attention mechanism over nodes in GNNs with experiments in a controlled environment. However, the attention mechanism discussed in the work is used in the pooling layer for the pooling of nodes, while our work investigates the usage of attention mechanism in the aggregation layer for the updating of nodes.

\section{Limitation of Attention-Based GNNs}
In this section, we theoretically analyze the discriminative power of attention-based GNNs and show their limitations. The discriminative power means how well an attention-based GNN can distinguish different elements (local or global structures). We find that previously proposed attention-based GNNs can fail in certain cases and the discriminative power is limited. Besides, by theoretically finding out all cases that always fail an attention-based GNN, we reveal that those failures come from the lack of cardinality preservation in attention-based aggregators. The details of proofs are included in the Supplemental Material.

\subsection{Discriminative Power of Attention-based GNNs}

We assume the node input feature space is countable. For any attention-based GNNs, we give the conditions in Lemma \ref{lemma:condition} to make them reach the upper bound of discriminative power when distinguishing different elements (local or global structures). In particular, each local structure belongs to a node and is the $k$-height subtree structure rooted at the node, which is naturally captured in the node feature $h _ { i } ^ { k }$ after $k$ iterations in a GNN. The global structure contains the information of all such subtrees in a graph.

\begin{lemma} \label{lemma:condition}
Let $\mathcal{A}: \mathcal{G} \rightarrow \mathbb{R}^g$ be a GNN following the neighborhood aggregation scheme with the attention-based aggregator (Equation \ref{weight-sum}). For global-level task, an extra readout function (Equation \ref{READOUT}) is used in the final layer. $\mathcal{A}$ can reach its upper bound of discriminative power (can distinguish all distinct local structures or be as powerful as the 1-WL test when distinguishing distinct global structures) after sufficient iterations with the following conditions:
\begin{itemize}
\item $\textbf{Local-level}$: Function $f$ and the weighted summation in Equation \ref{weight-sum} are injective.
\item $\textbf{Global-level}$: Besides the conditions for local-level, $\mathcal{A}$'s readout function (Equation \ref{READOUT}) is injective.
\end{itemize}
\end{lemma}

With Lemma \ref{lemma:condition}, we are interested in whether its conditions can always be satisfied, so as to reach the upper bound of discriminative capacity of an attention-based GNN. Since the function $f$ and the global-level readout function can be predetermined to be injective, we focus on whether the weighted summation function in attention-based aggregator can be injective.

\subsection{The Non-Injectivity of Attention-Based Aggregator}\label{Examples}

In this part, we aim to answer the following two questions:

\begin{question} \label{Q1}
Can the attention-based GNNs actually reach the upper bound of discriminative power? In other words, can the weighted summation function in an attention-based aggregator be injective?
\end{question}

\begin{question} \label{Q2}
If not, can we determine \textbf {all} of the cases that prevent any kind of weighted summation function being injective?
\end{question}

Given a countable feature space $\mathcal{H}$, a weighted summation function is a mapping $W: \mathcal{H} \rightarrow \mathbb{R}^n$. The exact $W$ is determined by the attention weights $\alpha$ computed from $Att$ in Equation \ref{attention}. Since $Att$ is affected by stochastic optimization algorithms (e.g. SGD) which introduce stochasticity in $W$, we have to pay attention that $W$ is not fixed when dealing with the two questions. 

In Theorem \ref{theorem:att}, we answer Q\ref{Q1} with {\em  No} by giving the cases that make $W$ not to be injective. So that the attention-based GNNs can \textbf {never} meet their upper bound of discriminative power, which is stated in Corollary \ref{corollary:WL}. Moreover, we answer Q\ref{Q2} with {\em  Yes} in Theorem \ref{theorem:att} by pointing out those cases are the \textbf {only} reason to always prevent $W$ being injective. This alleviates the difficulty of summarizing the properties of those cases. Besides, we can specifically propose methods to avoid those cases so as to let $W$ to be injective.

\begin{theorem} \label{theorem:att}
Assume the input feature space $\mathcal{X}$ is countable. Given a multiset $X \subset \mathcal{X}$ and the node feature $c$ of the central node, the weighted summation function $h (c, X)$ in aggregation is defined as $h (c, X) = \sum\nolimits _ { x \in X }\alpha _ { cx } f(x)$, where $f: \mathcal{X} \rightarrow \mathbb{R}^n$ is a mapping of input feature vector and $\alpha _ { cx }$ is the attention weight between $f(c)$ and $f(x)$ calculated by the attention function $Att$ in Equation \ref{attention} and the softmax function in Equation \ref{softmax}. For all $f$ and $Att$, $h(c_1, X_1) = h(c_2, X_2)$ \textbf {if and only if} $c_1 = c_2$, $X_1 = (S, \mu)$ and $X_2 = (S, k \cdot \mu)$ for $k \in \mathbb{N}^{*}$. In other words, $h$ will map different multisets to the same embedding if and only if the multisets have the same central node feature and the same distribution of node features.
\end{theorem}

\begin{corollary} \label{corollary:WL}
Let $\mathcal{A}$ be the GNN defined in Lemma \ref{lemma:condition}. $\mathcal{A}$ never reaches its upper bound of discriminative power:

There exists two different subtrees $S_1$ and $S_2$ or two graphs $G_1$ and $G_2$ that the Weisfeiler-Lehman test decides as non-isomorphic, such that $\mathcal{A}$ always maps the two subtrees/graphs to the same embeddings.
\end{corollary}

\subsection{Attention Mechanism Fails to Preserve Cardinality}\label{flaw}

With Theorem \ref{theorem:att}, we are now interested in the properties of all cases that always prevent the weighted summation functions $W$ being injective. Since the multisets that all $W$ fail to distinguish share the same distribution of node features, we can say that $W$ ignores the multiplicity information of each identical element in the multisets. Thus the cardinality of the multiset is not preserved:

\begin{corollary} \label{corollary:cardinality}
Let $\mathcal{A}$ be the GNN defined in Lemma \ref{lemma:condition}. The attention-based aggregator in $\mathcal{A}$ cannot preserve the cardinality information of the multiset of node features in aggregation.
\end{corollary}

In the next section, we aim to propose improved attention-based models to preserve the cardinality in aggregation.

\begin{figure}[t]
	\centering
	\includegraphics[scale=0.9]{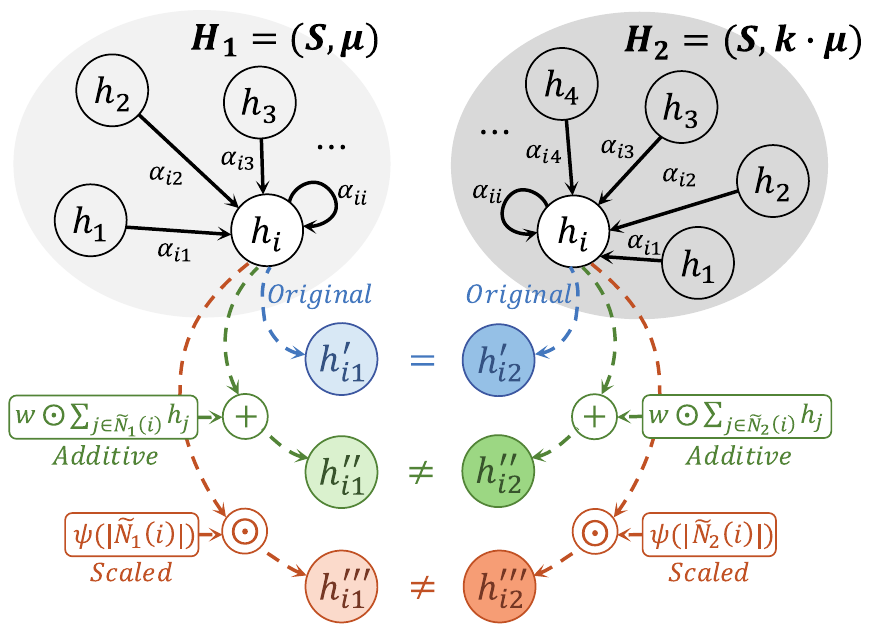}
	\caption{An illustration of different attention-based aggregators on multiset of node features. Given two distinct multisets $H_1$ and $H_2$ that have the same central node feature $h_i$ and the same distribution of node features, aggregators will map $h_i$ to $h_{i1}$ and $h_{i2}$ for $H_1$ and $H_2$. The \textit{Original} model will get $h'_{i1} = h'_{i2}$ and fail to distinguish $H_1$ and $H_2$, while our \textit{Additive} and \textit{Scaled} models can always distinguish $H_1$ and $H_2$ with $h''_{i1} \neq h''_{i2}$ and $h'''_{i1} \neq h'''_{i2}$.} \label{CPA}
\end{figure}

\section{Cardinality Preserved Attention (CPA) Model}

Since the cardinality of the multiset is not preserved in attention-based aggregators, our goal is to propose modifications to any kind of attention mechanism to make them capture the cardinality information. So that all of the cases that always prevent attention-based aggregator being injective can be avoided. 

To achieve our goal, we modify the weighted summation function in Equation \ref{weight-sum} to incorporate the cardinality information and don't change the attention function in Equation \ref{attention} so as to keep its original expressive power. Two different models named as \textit{Additive} and \textit{Scaled} are proposed to modify the \textit{Original} model in Equation \ref{weight-sum}:

\begin{model}\label{Additive-model}
(Additive)
\begin{align}
h _ { i } ^ { l } =  f ^ { l } \Big( \sum\nolimits _ { j \in \tilde{\mathcal { N }} ( i )}\alpha _ { i j } ^ { l-1 } h _ { j } ^ {  l-1 }+ w  ^ {  l} \odot \sum\nolimits _ { j \in \tilde{\mathcal { N }} ( i )} h _ { j } ^ {  l-1 } \Big), \label{Additive}
\end{align}
\end{model}
\begin{model}\label{Scaled-model}
(Scaled)
\begin{align}
h _ { i } ^ { l } =  f ^ { l } \Big( \psi^ {  l } \big(\big\vert \tilde{\mathcal { N }} ( i ) \big\vert\big) \odot \sum\nolimits _ { j \in \tilde{\mathcal { N }} ( i )}\alpha _ { i j } ^ { l-1 }  h _ { j } ^ {  l-1 } \Big), \label{Scaled}
\end{align}
where $w$ is a non-zero vector $\in \mathbb{R}^n$, $\odot$ denotes the element-wise multiplication, $\vert \tilde{\mathcal { N }} ( i ) \vert$ equals to the cardinality of the multiset $\tilde{\mathcal { N }} ( i )$, $\psi: \mathbb{Z}^+ \rightarrow \mathbb{R}^n$ is an injective function.
\end{model}

In the \textit{Additive} model, each element in the multiset will contribute to the term that we added to preserve the cardinality information. In the \textit{Scaled} model, the original weighted summation is directly multiplied by a representational vector of the cardinality value. So with these models, distinct multisets with the same distribution will result in different embedding $h$. Note that both of our models don't change the $Att$ function, such that they can keep the learning power of the original attention mechanism. We summarize the effect of our models in Corollary \ref{corollary:modified-att} and illustrate it in Figure \ref{CPA}.

\begin{corollary} \label{corollary:modified-att}
Let $\mathcal{T}$ be the original attention-based aggregator in Equation \ref{weight-sum}. With our proposed \textbf {Cardinality Preserved Attention (CPA)} models in Equation \ref{Additive} and \ref{Scaled}, $\mathcal{T}'s$ discriminative power is increased: $\mathcal{T}$ can now distinguish \textbf {all} different multisets in aggregation that it previously always fails to distinguish.
\end{corollary}

While the original attention-based aggregator is never injective as we mentioned in previous sections, our cardinality preserved attention-based aggregator can be injective with certain learned attention weights to reach its upper bound of discriminative power. We validate this in our experiments. 

For the time and space complexity of our CPA models comparing to the original attention-based aggregator, it is obvious that the Model \ref{Additive-model} and \ref{Scaled-model} take more time and space than the original one due to our introduced vectors $w$ and $\psi (\vert \tilde{\mathcal { N }} ( i ) \vert)$. Thus we further simplify our models by fixing the values in $w$ and $\psi (\vert \tilde{\mathcal { N }} ( i ) \vert)$ and define two CPA variants:

\begin{model}\label{f-Additive-model}
(f-Additive)
\begin{align}
h _ { i } ^ { l } =  f ^ { l }\Big(\sum\nolimits _ { j \in \tilde{\mathcal { N }} ( i )}(\alpha _ { i j } ^ { l-1 } + 1) h _ { j } ^ {  l-1 } \Big), \label{f-Additive}
\end{align}
\end{model}
\begin{model}\label{f-Scaled-model}
(f-Scaled)
\begin{align}
h _ { i } ^ { l } =  f ^ { l } \Big( \big\vert \tilde{\mathcal { N }} ( i ) \big\vert \cdot \sum\nolimits _ { j \in \tilde{\mathcal { N }} ( i )}\alpha _ { i j } ^ { l-1 }  h _ { j } ^ { l-1 } \Big). \label{f-Scaled}
\end{align}
\end{model}

Model \ref{f-Additive-model} and \ref{f-Scaled-model} still preserve the cardinality information and have reduced time and space complexity comparing to Model \ref{Additive-model} and \ref{Scaled-model}. Actually, since $w$ and $\psi (\vert \tilde{\mathcal { N }} ( i ) \vert)$ are degenerate into constants, Model \ref{f-Additive-model} and \ref{f-Scaled-model} have the \textbf {same} time and space complexity as the original model in Equation \ref{weight-sum}. In our experiments, we will examine all 4 models together with the original one.

\section{Experiments}
In our experiments, we focus on the following questions:

\begin{question} \label{Q3}
Since attention-based GNNs (e.g. GAT) are originally proposed for local-level tasks like node classification, will those models fail or not meet the upper bound of discriminative power when solving certain node classification tasks? If so, can our proposed CPA models improve the original model?
\end{question}

\begin{question} \label{Q4}
For global-level tasks like graph classification, how well can the original attention-based GNNs perform? Can our proposed CPA models improve the original model?
\end{question}

\begin{question} \label{Q5}
How the attention-based GNNs with our CPA models perform compared to baselines?
\end{question}

To answer Question \ref{Q3}, we design a node classification task which is to predict whether or not a node is included in a triangle as one vertex in a graph. To answer Question \ref{Q4} and \ref{Q5}, we perform experiments on graph classification benchmarks and evaluate the performance of attention-based GNNs with CPA models.

\subsection{Experimental Setup}

\subsubsection{Datasets}
In our synthetic task (TRIANGLE-NODE) for predicting whether or not a node is included in a triangle, we generate a graph with different node features. In our experiment on graph classification, we use 6 benchmark datasets: 2 social network datasets (REDDIT-BINARY (RE-B), REDDIT-MULTI5K (RE-M5K)) and 4 bioinformatics datasets (MUTAG, PROTEINS, ENZYMES, NCI1). More details of the datasets are provided in Supplemental Material. 

\begin{table}[t]
\centering
\small
\caption{Testing accuracies(\%) of GAT variants (the original GAT and the GAT applied with each of our 4 CPA models) on TRIANGLE-NODE dataset for node classification. We highlight the result of the best performed model. The proportion $P$ of multisets that hold the properties in Theorem \ref{theorem:att} among all multisets is also reported.} \label{Triangle}
\smallskip
\begin{tabular}{cc}
\toprule
Dataset & {TRIANGLE-NODE}\\
$P(\%)$  & 29.2 \\
\midrule
\midrule
{Original}   &  78.40 $\pm$ 7.65\\
\midrule
{Additive}   &  91.31 $\pm$ 1.19\\
{Scaled}     &  {\bf 91.38 $\pm$ 1.23}\\
{f-Additive} &  91.18 $\pm$ 1.24\\
{f-Scaled}   &  91.36 $\pm$ 1.26\\
\bottomrule
\end{tabular}
\end{table}

\begin{table}[t]
\centering
\small
\caption{Testing accuracies(\%) of GAT-GC variants (the original one and the ones applied with each of our 4 CPA models) on social network datasets. We highlight the result of the best performed model per dataset. The proportion $P$ of multisets that hold the properties in Theorem \ref{theorem:att} among all multisets is also reported for each dataset.} \label{Social}
\smallskip
\begin{tabular}{ccc}
\toprule
Datasets &{RE-B}&{RE-M5K}\\
$P(\%)$  & 100.0 &  100.0 \\
\midrule
\midrule
{Original}  & 50.00 $\pm$ 0.00   & 20.00 $\pm$ 0.00 \\
\midrule
{Additive}  & {\bf 93.07 $\pm$ 1.82} & {\bf 57.39 $\pm$ 2.09} \\
{Scaled}  &  92.36 $\pm$ 2.27  &  56.76 $\pm$ 2.26 \\
{f-Additive}  & 93.05 $\pm$ 1.87   &  56.43 $\pm$ 2.38 \\
{f-Scaled}  &  92.57 $\pm$ 2.06 &  57.22 $\pm$ 2.20 \\
\bottomrule
\end{tabular}
\end{table}

\begin{figure}[t]
	\centering
	\includegraphics[scale=0.2]{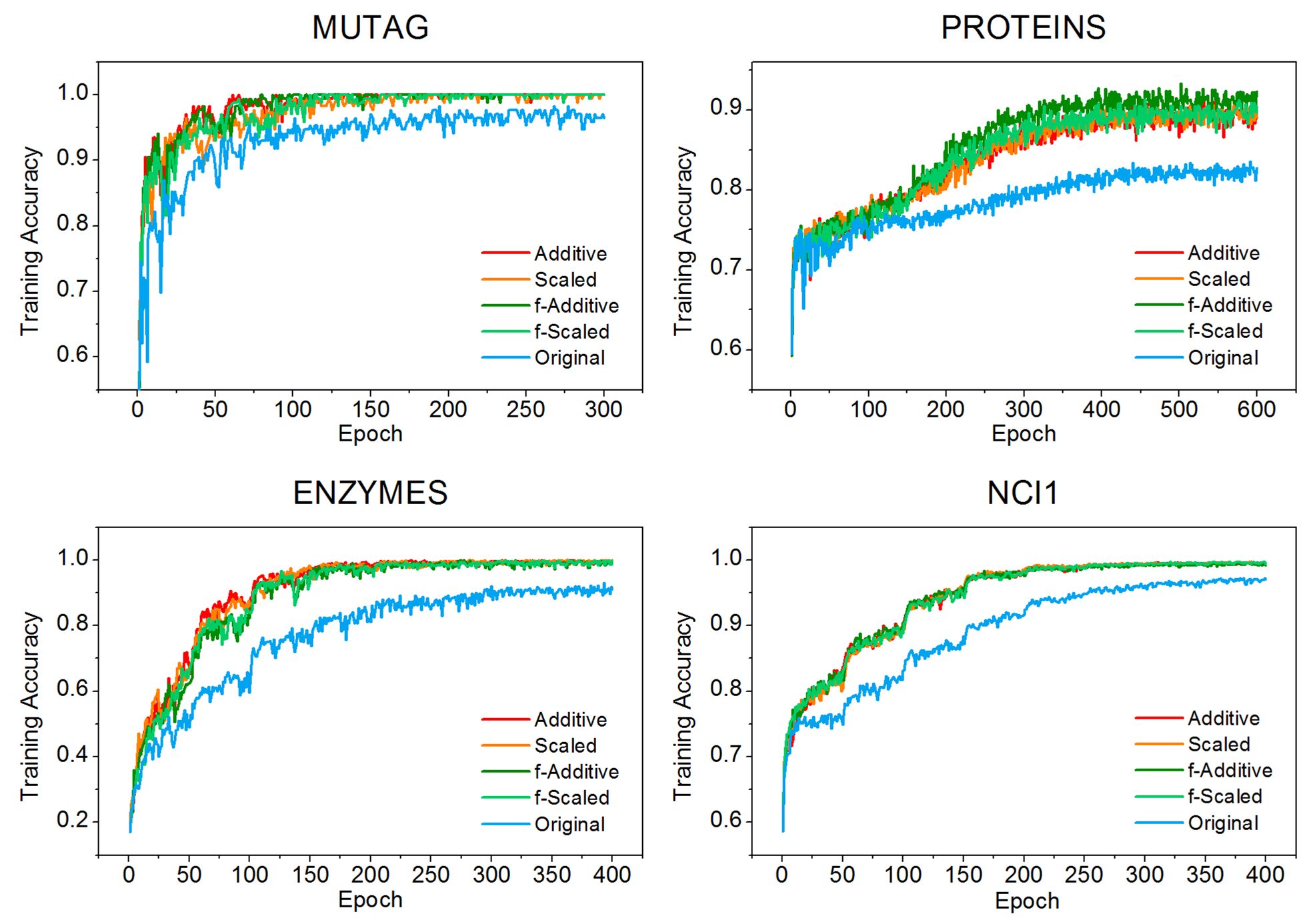}
	\caption{Training curves of GAT-GC variants on bioinformatics datasets.} \label{Training_curve}
\end{figure}

\begin{table}[t]
\centering
\scriptsize
\caption{Testing accuracies(\%) of GAT-GC variants (the original one and the ones applied with each of our 4 CPA models) on bioinformatics datasets. We highlight the result of the best performed model per dataset. The highlighted results are significantly higher than those from the corresponding \textit{Original} model under paired t-test at significance level $5\%$. The proportion $P$ of multisets that hold the properties in Theorem \ref{theorem:att} among all multisets is also reported for each dataset.} \label{Bioinfo}
\smallskip
\begin{tabular}{ccccc}
\toprule
{Datasets}&{MUTAG}&{PROTEINS}&{ENZYMES}&{NCI1}\\
$P(\%)$ & 56.9 & 29.3 & 29.4 & 43.3 \\
\midrule
\midrule
{Original}  & 84.96 $\pm$ 7.65   & 75.64 $\pm$ 3.96   &  58.08 $\pm$ 6.82 &  80.29 $\pm$ 1.89\\
\midrule
{Additive}  & 89.75 $\pm$ 6.39   &  76.61 $\pm$ 3.80  &  58.90 $\pm$ 6.96 &  81.92 $\pm$ 1.89\\
{Scaled}  &  89.65 $\pm$ 7.47  &  76.44 $\pm$ 3.77   &  58.35 $\pm$ 6.97 &  82.18 $\pm$ 1.67\\
{f-Additive} & 90.34 $\pm$ 6.05  &  76.60 $\pm$ 3.91  &  {\bf 59.80 $\pm$ 6.18} &  81.96 $\pm$ 2.01\\
{f-Scaled}  & {\bf 90.44 $\pm$ 6.44}  &  {\bf 76.81 $\pm$ 3.77}   &  58.45 $\pm$ 6.35 &  {\bf 82.28 $\pm$ 1.81} \\
\bottomrule
\end{tabular}
\end{table}

\begin{table*}[t]
\centering
\small
\caption{Testing accuracies(\%) for graph classification. We highlight the result of the best performed model for each dataset. Our GAT-GC (f-Scaled) model achieves the top 2 on all 6 datasets.} \label{Testing_results}
\smallskip
\begin{tabular}{cccccccc}
\toprule
&Datasets  &{MUTAG}&{PROTEINS}&{ENZYMES}&{NCI1}&{RE-B}&{RE-M5K}\\
\midrule
\midrule
\multirow{5}{*}{\rotatebox{90}{\hspace*{-1pt}Baselines}}    
&{WL} & 82.05 $\pm$ 0.36   & 74.68 $\pm$ 0.49   & 52.22 $\pm$ 1.26 &  82.19 $\pm$ 0.18  & 81.10 $\pm$ 1.90  & 49.44 $\pm$ 2.36\\
&{PSCN} & 88.95 $\pm$ 4.37  & 75.00 $\pm$ 2.51  & -   &  76.34 $\pm$ 1.68 & 86.30 $\pm$ 1.58  & 49.10 $\pm$ 0.70  \\
&{DGCNN}& 85.83 $\pm$ 1.66   & 75.54 $\pm$ 0.94  & 51.00 $\pm$ 7.29  & 74.44 $\pm$ 0.47  & 76.02 $\pm$ 1.73  & 48.70 $\pm$ 4.54  \\
&{GIN}  &  89.40 $\pm$ 5.60     & 76.20 $\pm$ 2.80   & -  & {\bf 82.70 $\pm$ 1.70}  & 92.40 $\pm$ 2.50    & {\bf 57.50 $\pm$ 1.50} \\
&{CapsGNN}  &  86.67 $\pm$ 6.88     & 76.28 $\pm$ 3.63   & 54.67 $\pm$ 5.67  & 78.35 $\pm$ 1.55  & -    & 52.88 $\pm$ 1.48 \\
\midrule
&{GAT-GC (f-Scaled)}  & {\bf 90.44 $\pm$ 6.44}   & {\bf 76.81 $\pm$ 3.77}   &  {\bf 58.45 $\pm$ 6.35} &  82.28 $\pm$ 1.81 &  {\bf 92.57 $\pm$ 2.06} & 57.22 $\pm$ 2.20 \\
\bottomrule
\end{tabular}
\end{table*}

\subsubsection{Models}
In our experiments, the \textbf {\textit{Original}} model is the one that uses the original version of an attention mechanism. We apply each of our 4 CPA models (\textbf {\textit{Additive}}, \textbf {\textit{Scaled}}, \textbf {\textit{f-Additive}} and \textbf {\textit{f-Scaled}}) to the original attention mechanism for comparison. In the \textit{Additive} and \textit{Scaled} models, we take advantage of the approximation capability of multi-layer perceptron (MLP)~\cite{hornik1989multilayer,hornik1991approximation} to model $f$ and $\psi$.

For node classification, we use GAT~\cite{velickovic2018graph} as the \textit{Original} model. For graph classification, we build a GNN (GAT-GC) based on GAT as the \textit{Original} model:  We adopt the attention mechanism in GAT to specify the form of Equation \ref{attention}: $e _ { i j } =  \operatorname { LeakyReLU } \left( \mathbf { a } ^{\top} \left[ \mathbf { W }  { h } _ { i } \| \mathbf { W } { h } _ { j } \right] \right)$. For the readout function, a naive way is to only consider the node embeddings from the last iteration. Although a sufficient number of iterations can help to avoid the cases in Theorem \ref{theorem:att} by aggregating more diverse node features, the features from the latter iterations may generalize worse and the GNNs usually have shallow structures~\cite{xu2018how,zhou2018graph}. So the GAT-GC adopts the same function as used in~\cite{xu2018representation,xu2018how,lee2019self,li2019deepgcns}, which concatenates graph embeddings from all iterations: $h_{G} = \concat_{k=0}^L \big( \operatorname{ Readout }( \left\{ h _ { i } ^ { l } \big\vert i \in G \right\} ) \big)$, $\operatorname{ Readout }$ function can be sum or mean. With CPA models, the cases in Theorem \ref{theorem:att} can be avoided in each iteration. Full experimental settings are included in Supplemental Material. 

\subsection{Node Classification}
For the TRIANGLE-NODE dataset, the proportion P of multisets that hold the properties in Theorem \ref{theorem:att} is $29.2\%$, as shown in Table \ref{Triangle}. The classification accuracy of the \textit{Original} model (GAT) is significantly lower than the CPA models. It supports the claim in Corollary \ref{corollary:WL}: the \textit{Original} model fails to distinguish all distinct multisets in the dataset and exhibits constrained discriminate power. On the contrary, CPA models can distinguish all different multisets in the graph as suggested in Corollary \ref{corollary:modified-att} and indeed significantly improve the accuracy of the \textit{Original} model as shown in Table \ref{Triangle}. This experiment thus well answers Question \ref{Q3} that we raised.

\subsection{Graph Classification}

In this section, we aim to answer Question \ref{Q4} by evaluating the performance of variants of GAT-based GNN (GAT-GC) on graph classification benchmarks. Besides, we compare our best-performed CPA model with baseline models to answer Question \ref{Q5}.

\subsubsection{Social Network Datasets}
Since the RE-B and RE-M5K datasets don't have original node features and we assign all the node features to be the same, we have $P=100.0\%$ in those datasets. Thus \textbf {all} multisets in aggregation will be mapped to the same embedding by the \textit{Original} GAT-GC. After a mean readout function on all multisets, \textbf {all} graphs are finally mapped to the same embedding. The performance of the \textit{Original} model is just random guessing of the graph labels as reported in Table \ref{Social}. While our CPA models can distinguish all different multisets and are confirmed to be significantly better than the \textit{Original} one. 

Here we examine a naive approach to incorporate the cardinality information in the \textit{Original} model by assigning node degrees as input node labels. By doing this way, the node features are diverse and we get $P=0.0\%$, which means that the cases in Theorem \ref{theorem:att} can be all avoided. However, the testing accuracies of \textit{Original} can only reach $76.65 \pm 9.87\%$ on \textsc{RE-B} and $43.71 \pm 9.05\%$ on \textsc{RE-M5K}, which are significantly lower than the results of CPA models in Table \ref{Social}. Thus in practice, our proposed models exhibit good generalization power comparing to the naive approach.

\subsubsection{Bioinformatics Datasets}
For bioinformatics datasets that contain diverse node labels, we also report the $P$ values in Table \ref{Bioinfo}. The results reveal the existence ($P \geq 29.3\%$) of the cases in those datasets that can fool the \textit{Original} model, thus the discriminative power of the \textit{Original} model is {\em theoretically} constrained. 

To empirically validate this, we compare the training accuracies of GAT-GC variants, since the discriminative power can be directly indicated by the accuracies on {\em training sets}. Higher training accuracy indicates a better fitting ability to distinguish different graphs. The training curves of GAT-GC variants are shown in Figure \ref{Training_curve}. From these curves, we can see even though the \textit{Original} model has overfitted different datasets, the fitting accuracies that it converges to can never be higher than those of our CPA models. Compared to the WL kernel, CPA models can get training accuracies close to $100\%$ on several datasets, which reach those obtained from the WL kernel (equal to $100\%$ as shown in~\cite{xu2018how}). These findings validate that the discriminative power of the \textit{Original} model is constrained while our CPA models can approach the upper bound of discriminative power with certain learned weights.

In Table \ref{Bioinfo} we report the testing accuracies of GAT-GC variants on bioinformatics datasets. The \textit{Original} model can get meaningful results. However, we find our proposed CPA models further improve the testing accuracies of the \textit{Original} model on all datasets. This indicates that the preservation of cardinality can also benefit the generalization power of the model besides the discriminative power.

From previous results in Table \ref{Social} and \ref{Bioinfo}, we find the \textit{f-Scaled} model performs the best with an average ranking measure~\cite{taheri2018learning}. The good performance of the fixed-weight models (\textit{f-Additive} and \textit{f-Scaled}) comparing to the full models (\textit{Additive} and \textit{Scaled}) demonstrates that the improvements achieved by CPA models are not simply due to the increased capacities given by the additional vectors embedded.

\subsubsection{Comparison to Baselines}
We further compare the best-performed GAT-GC variant (\textit{f-Scaled}) with other baselines (WL kernel (WL)~\cite{shervashidze2011weisfeiler}, PATCHY-SAN (PSCN)~\cite{niepert2016learning}, Deep Graph CNN (DGCNN)~\cite{zhang2018end}, Graph Isomorphism Network (GIN)~\cite{xu2018how} and Capsule Graph Neural Network (CapsGNN)~\cite{xinyi2019capsule}). In Table \ref{Testing_results}, we report the results. Our GAT-GC (f-Scaled) model achieves 4 top 1 and 2 top 2 on all 6 datasets. It is expected that even better performance can be achieved with certain choices of attention mechanism besides the GAT one.

\section{Conclusion}
In this paper, we theoretically analyze the representational power of GNNs with attention-based aggregators: We determine all cases when those GNNs always fail to distinguish distinct structures. The finding shows that the missing cardinality information in aggregation is the only reason to cause those failures. To improve, we propose cardinality preserved attention (CPA) models to solve this issue. In our experiments, we validate our theoretical analysis that the performances of the original attention-based GNNs are limited. With our models, the original models can be improved. Compared to other baselines, our best-performed model achieves competitive performance. In future work, a challenging problem is to better learn the attention weights so as to guarantee the injectivity of our cardinality preserved attention models after the training. Besides, it would be interesting to analyze the effects of different attention mechanisms.

\bibliographystyle{aaai}
\bibliography{main}

\newpage
\appendix
\section{Proof for Lemma \ref{lemma:condition}}
\begin{proof} Local-level: For the aggregator in the first layer, it will map different 1-height subtree structures to different embeddings from the distinct input multisets of neighborhood node features, since it's injective. Iteratively, the aggregator in the $l$-th layer can distinguish different $l$-height subtree structures by mapping them to different embeddings from the distinct input multisets of $l$-1-height subtree features, since it's injective.

Global level: From Lemma 2 and Theorem 3 in~\cite{xu2018how}, we know: When all functions in $\mathcal{A}$ are injective, $\mathcal{A}$ can reach its upper bound of discriminative power, which is the same as the Weisfeiler-Lehman (WL) test~\cite{weisfeiler1968reduction} when deciding the graph isomorphism. 
\end{proof}

\section{Proof for Theorem \ref{theorem:att}}
\begin{proof}
To prove Theorem \ref{theorem:att}, we have consider both two directions in the iff statement:

\paragraph{(1)} If given $c_1 = c_2 = c$, $X_1 = (S, \mu)$ and $X_2 = (S, k \cdot \mu)$, as $h (c, X) = \sum _ { x \in X }\alpha _ { cx } f(x)$, we have:
\[h (c_i, X_i) = \sum _ { x \in X_i }\alpha _ { cxi } f(x), i\in\{1,2\},\]
where $\alpha _ { cxi}$ is the attention weight belongs to $X_i$, and between $f(c)$ and $f(x)$, $x \in X_i, i\in\{1,2\}$.

We can rewrite the equations using $S$ and $\mu$: 
\[h (c_1, X_1) = h (c, S, \mu) = \sum_{s\in S}\mu(s)\alpha_{cs1} f(s),\]
\[h (c_2, X_2) = h (c, S, k \cdot \mu) = \sum_{s\in S}k \cdot \mu(s) \alpha _ {cs2} f(s),\]
where $\mu(s)$ is the multiplicity function, and $\alpha _ { csi}$ is the attention weight belongs to $X_i$, and between $f(c)$ and $f(s)$, $s \in S, i\in\{1,2\}$.

Considering the softmax function in Equation 2 of our paper, we can use attention coefficient $e$ to rewrite the equations:
\begin{align}
\sum_{s\in S}\mu(s)\alpha_{cs1} f(s) &= \sum_{s\in S}\mu(s)\frac{\exp (e_{cs1})}{\sum _ { x \in X_1 }\exp(e_{cx1})} f(s) \nonumber\\
&= \frac{\sum_{s\in S}\mu(s) \exp (e_{cs1})}{\sum _ { x \in X_1 }\exp(e_{cx1})} f(s),\nonumber\\
\sum_{s\in S}k \cdot \mu(s) \alpha _ {cs2} f(s) &= k \cdot \sum_{s\in S}\mu(s)\frac{\exp (e_{cs2})}{\sum _ { x \in X_2 }\exp(e_{cx2})} f(s) \nonumber\\
&= k \cdot \frac{\sum_{s\in S}\mu(s) \exp (e_{cs2})}{\sum _ { x \in X_2 }\exp(e_{cx2})} f(s), \nonumber
\end{align}
where $e _ { csi }$ is the attention coefficient belongs to $X_i$, and between $f(c)$ and $f(s)$, $s \in S, i\in\{1,2\}$. Moreover, $e _ { cxi }$ is the attention coefficient belongs to $X_i$, and between $f(c)$ and $f(x)$, $x \in X_i, i\in\{1,2\}$.

As attention coefficient $e$ is computed by function $Att$, which is regardless of $X$, thus $e _ { cs1 }=e _ { cs2 }$, $\forall s \in S$ and $e _ { cx1 }=e _ { cx2 }$, $\forall x \in X_1, X_2$. We denote $e _ { cx }=e _ { cx1 }=e _ { cx2 }$, $e _ { cs }=e _ { cs1 }=e _ { cs2 }$. Remind that $X_2$ has $k$ copies of the elements in $X_1$, so that
\[ \sum _ { x \in X_1 }\exp(e_{cx}) = \frac{1}{k} \sum _ { x \in X_2 }\exp(e_{cx}).\]

Using this equation, we can get
\begin{align}
\frac{\sum_{s\in S}\mu(s) \exp (e_{cs1})}{\sum _ { x \in X_1 }\exp(e_{cx1})} f(s)& = \frac{\sum_{s\in S}\mu(s) \exp (e_{cs})}{\frac{1}{k} \sum _ { x \in X_2 }\exp(e_{cx})} f(s)\nonumber\\
&= k \cdot \frac{\sum_{s\in S}\mu(s) \exp (e_{cs2})}{\sum _ { x \in X_2 }\exp(e_{cx2})} f(s) .\nonumber
\end{align}

From all equations above, we finally have
\begin{align}
h (c_1, X_1) &= \frac{\sum_{s\in S}\mu(s) \exp (e_{cs1})}{\sum _ { x \in X_1 }\exp(e_{cx1})} f(s) \nonumber\\
&= k \cdot \frac{\sum_{s\in S}\mu(s) \exp (e_{cs2})}{\sum _ { x \in X_2 }\exp(e_{cx2})} f(s) \nonumber\\
&= h (c_2, X_2). \nonumber
\end{align}

\paragraph{(2)} If given $h(c_1, X_1) = h(c_2, X_2)$ for all $f$, $Att$, we have 
\[\sum _ { x \in X_1 }\alpha _ { cx1 } f(x)=\sum _ { x \in X_2 }\alpha _ { cx2 } f(x), \quad \forall f, Att,\]
where $\alpha _ { cxi }$ is the attention weight belongs to $X_i$, and between $f(c_i)$ and $f(x)$, $x \in X_i, i\in\{1,2\}$.

We denote $X_1 = (S_1, \mu_1)$ and $X_2 = (S_2, \mu_2)$ and rewrite the equation: 
\[\sum_{s\in S_1}\mu_{1}(s)\alpha_{cs1} f(s)=\sum_{s_\in S_2}\mu_{2}(s)\alpha_{cs2} f(s), \quad \forall f, Att,\]
where $\mu_{i}(s)$ is the multiplicity function of $X_i, i\in\{1,2\}$. Moreover, $\alpha _ { csi }$ is the attention weight belongs to $X_i$, and between $f(c_i)$ and $f(s)$, $s \in S_i, i\in\{1,2\}$. 

When considering the relations between $S_1$ and $S_2$, we have: 
\begin{align}
\sum_{s\in S_1\cap S_2}&\big(\mu_{1}(s)\alpha_{cs1} - \mu_{2}(s) \alpha_{cs2}\big) f(s)+\nonumber\\
\sum_{s\in S_1 \setminus S_2}&\mu_{1}(s)\alpha_{cs1} f(s)-\sum_{s\in S_2 \setminus S_1} \mu_{2}(s) \alpha_{cs2} f(s)=0.\label{Equation1}
\end{align}

If we assume the equality of Equation \ref{Equation1} is true for all $f$ and $S_1\neq S_2$, we can define such two functions $f_1$ and $f_2$: 
\begin{align}
&f_1(s)=f_2(s), \ \ \ \ \ \ \ \ \ \forall s\in S_1\cap S_2,\nonumber\\
&f_1(s)=f_2(s)-1, \ \ \ \forall s\in S_1\setminus S_2,\nonumber\\ &f_1(s)=f_2(s)+1, \ \ \ \forall s\in S_2\setminus S_1.\nonumber
\end{align}

If given the equality of Equation \ref{Equation1} is true for $f_1$, we have:
\begin{align}
\sum_{s\in S_1\cap S_2}&\big(\mu_{1}(s)\alpha_{cs1} - \mu_{2}(s) \alpha_{cs2}\big) f_1(s)+\nonumber\\
\sum_{s\in S_1 \setminus S_2}&\mu_{1}(s)\alpha_{cs1} f_1(s)-\sum_{s\in S_2 \setminus S_1} \mu_{2}(s) \alpha_{cs2} f_1(s)=0.\label{Equation_f1}
\end{align}

We can rewrite Equation \ref{Equation_f1} using $f_2$:
\begin{align}
&\sum_{s\in S_1\cap S_2}\big(\mu_{1}(s)\alpha_{cs1} - \mu_{2}(s) \alpha_{cs2}\big) f_2(s)+\nonumber\\
&\sum_{s\in S_1 \setminus S_2}\mu_{1}(s)\alpha_{cs1} (f_2(s)-1)-\nonumber\\
&\sum_{s\in S_2 \setminus S_1} \mu_{2}(s) \alpha_{cs2} (f_2(s)+1)=0.\nonumber
\end{align}

Thus we know
\begin{align}
&\sum_{s\in S_1\cap S_2}\big(\mu_{1}(s)\alpha_{cs1} - \mu_{2}(s) \alpha_{cs2}\big) f_2(s)+\nonumber\\
&\sum_{s\in S_1 \setminus S_2}\mu_{1}(s)\alpha_{cs1} f_2(s)-\sum_{s\in S_2 \setminus S_1} \mu_{2}(s) \alpha_{cs2} f_2(s)= \nonumber\\
&\sum_{s\in S_1 \setminus S_2}\mu_{1}(s)\alpha_{cs1}+\sum_{s\in S_2 \setminus S_1} \mu_{2}(s) \alpha_{cs2}
\label{Equation_f2}
\end{align}

Note that the LHS of Equation \ref{Equation_f2} is just the LHS of Equation \ref{Equation1} when $f=f_2$. As $\mu_i(s)\geq 1$ due to the definition of multiplicity, $\alpha_{csi}>0$ due to the softmax function, we have $\mu_i(s) \alpha_{csi}>0, \forall s\in S_i, i\in\{1,2\}$. Thus the RHS of Equation \ref{Equation_f2} > 0 and we now know the equality in Equation \ref{Equation1} is not true for $f_2$. So the assumption of $S_1\neq S_2$ is false. 

We denote $S=S_1=S_2$. To let the remaining summation term always equal to 0, we have 
\begin{align}
\mu_1(s)\alpha_{cs1}-\mu_2(s) \alpha_{cs2}=0, \quad \forall Att.\nonumber
\end{align}

Considering Equation 2 in our paper, we can rewrite the equation above: 
\begin{align}
\frac{\mu_1(s)}{\mu_2(s)} = \frac{\exp (e_{cs2})}{\exp (e_{cs1})} \frac{\sum _ { x \in X_1 }\exp(e_{cx1})}{\sum _ { x \in X_2 }\exp(e_{cx2})}, \quad \forall Att,\label{Equation2}
\end{align}
where $e _ { csi }$ is the attention coefficient belongs to $X_i$, and between $f(c_i)$ and $f(s)$, $s \in S$. And $e _ { cxi }$ is the attention coefficient belongs to $X_i$, and between $f(c_i)$ and $f(x)$, $x \in X_i, i\in\{1,2\}$.

The LHS of Equation \ref{Equation2} is a rational number. However if $c_1 \neq c_2$, the RHS of Equation \ref{Equation2} can be irrational: We assume $S$ contains at least two elements $s_0$ and $s \neq s_0$. If not, we can directly get $c_1=c_2$. We consider any attention mechanism that results in: 
\begin{align}
&e_{cs1}=1, \ \ \forall s \in S,\nonumber\\
&e_{cs2}=\left\{\begin{matrix}
1, & \textrm{for} s=s_0,\\ 
2, & \forall s \neq s_0 \in S.
\end{matrix}\right.\nonumber
\end{align}

Thus when $s=s_0$, the RHS of the equation become:
\begin{align}
\frac{e}{e}\frac{\big\vert X_1 \big\vert e}{(\big\vert X_2 \big\vert-n) e^2+ne}=\frac{\big\vert X_1 \big\vert}{(\big\vert X_2 \big\vert-n)e+n},\nonumber 
\end{align}
where $n$ is the multiplicity of $s_0$ in $X_2$. It is obvious that the value of RHS is irrational. So we have $c_1=c_2$ to always hold the equality.

With $c_1 = c_2$, we know $e _ { cs1 }=e _ { cs2 }$, $\forall s \in S$ and $e _ { cx1 }=e _ { cx2 }$, $\forall x \in X_1, X_2$. We denote $e _ { cx }=e _ { cx1 }=e _ { cx2 }$, Equation \ref{Equation2} becomes
\begin{align}
\frac{\mu_1(s)}{\mu_2(s)} = \frac{\sum _ { x \in X_1 }\exp(e_{cx})}{\sum _ { x \in X_2 }\exp(e_{cx})} = const., \quad \forall Att.\nonumber
\end{align}

We further denote $k = \mu_1(s)/\mu_2(s), \forall s \in S$. So that $\mu_2 = k \cdot \mu_1$. Finally by denoting $\mu = \mu_1$, we have $X_1 = (S, \mu)$, $X_2 = (S, k \cdot \mu)$ and $c_1 = c_2$.
\end{proof}

\section{Proof for Corollary \ref{corollary:WL}}
\begin{proof}
For subtrees, if $S_1$ and $S_2$ are 1-height subtrees that have the same root node feature and the same distribution of node features, $\mathcal{A}$ will get the same embeddings for $S_1$ and $S_2$ according to Theorem \ref{theorem:att}.

For graphs, let $G_1$ be a fully connect graph with $n$ nodes and $G_2$ be a ring-like graph with $n$ nodes. All nodes in $G_1$ and $G_2$ have the same feature $x$. It is clear that the Weisfeiler-Lehman test of isomorphism decides $G_1$ and $G_2$ as non-isomorphic.

We denote $\{X_i\}, i \in G_1$ as the set of multisets for aggregation in $G_1$, and $\{X_j\}, j \in G_2$ as the set of multisets for aggregation in $G_2$. As $G_1$ is a fully connect graph, all multisets in $G_1$ contain $1$ central node and $n-1$ neighbors. As $G_2$ is a ring-like graph, all multisets in $G_2$ contain $1$ central node and $2$ neighbors. Thus we have
\[X_i = (\{x\}, \{\mu_1(x) = n\}), \ \ \forall i \in G_1,\]
\[X_j = (\{x\}, \{\mu_2(x) = 3\}), \ \ \forall j \in G_2,\]
where $\mu_i(x)$ is the multiplicity function of the node with feature $x$ in $G_i, i\in\{1,2\}$.

From Theorem 1, we know that $h(c, X_i) = h(c, X_j), \forall i \in G_1, \forall j \in G_2$. Considering the Equation 3 of our paper, we have $h_i^l = h_j^l, \forall i \in G_1, \forall j \in G_2$ in each iteration $l$. Besides, as the number of node in $G_1$ and $G_2$ are equals to $n$, $\mathcal{A}$ will always map $G_1$ and $G_2$ to the same set of multisets of node features $\{h^l\}$ in each iteration $l$ and finally get the same embedding for each graph.
\end{proof}

\section{Proof for Corollary \ref{corollary:cardinality}}
\begin{proof}
Given two distinct multiset of node features $X_1$ and $X_2$ that have the same central node feature and the same distribution of node features: $c_1 = c_2$, $X_1 = (S, \mu)$ and $X_2 = (S, k \cdot \mu)$ for $k \in \mathbb{N}^{*}$, we know the cardinality of $X_2$ is $k$ times of the cardinality of $X_1$. Thus $X_1$ and $X_2$ can be distinguished by their cardinality. 

However, the weighted summation function $h$ in attention-based aggregator $\mathcal{A}$ will map them to the same embedding: $h(c_1, X_1) = h(c_2, X_2)$ according to Theorem \ref{theorem:att}. Thus we cannot distinguish $X_1$ and $X_2$ via $\mathcal{A}$. To conclude, $\mathcal{A}$ lost the cardinality information after aggregation.
\end{proof}

\section{Proof for Corollary \ref{corollary:modified-att}}
\begin{proof}
For any two distinct multisets $X_1$ and $X_2$ that $\mathcal{T}$ previously always fail to distinguish according to Theorem \ref{theorem:att}, we denote $X_1 = (S, \mu)$ and $X_2 = (S, k \cdot \mu) \subset \mathcal{X}$ for some $k \in \mathbb{N}^{*}$ and $c\in S$. Thus $\sum\nolimits _ { x \in X_1 }\alpha _ { cx1 } f(x)=\sum\nolimits _ { x \in X_2 }\alpha _ { cx2 } f(x)$, where $\alpha _ { cxi }$ is the attention weight belongs to $X_i$, and between $f(c)$ and $f(x)$, $x \in X_i, i\in\{1,2\}$. We denote $H = \sum _ { x \in X_1 }\alpha _ { cx1 } f(x)=\sum _ { x \in X_2 }\alpha _ { cx2 } f(x)$. When applying CPA models, the aggregation functions in $\mathcal{T}$ can be rewritten as:
\begin{align}
h_1(c, X_i) &= H + w  \odot \sum\nolimits _ { x \in X_i} f(x),\quad i\in\{1,2\},\nonumber \\
h_2(c, X_i) &= \psi (\big\vert X_i \big\vert) \odot H,\quad\quad\quad\quad\quad\  i\in\{1,2\}.\nonumber
\end{align}

We consider the following example: All elements in $w$ equal to 1. Function $\psi$ maps $\big\vert X \big\vert$ to a n-dimensional vector which all elements in it equal to $\big\vert X \big\vert$. And $f(x) = N^{-Z(x)}$, where $Z: \mathcal{X} \rightarrow \mathbb{N}$ and $N > \big\vert X \big\vert$. So that the aggregation functions become:
\begin{align}
h_1(c, X_i) &= H + \sum\nolimits _ { x \in X_i} f(x),\quad i\in\{1,2\},\nonumber \\
h_2(c, X_i) &= \big\vert X_i \big\vert \cdot H,\quad\quad\quad\quad\quad\  i\in\{1,2\}.\nonumber
\end{align}

For $h_1$, we have $h_1(c, X_1) - h_1(c, X_2) = \sum\nolimits _ { x \in X_1} f(x) - \sum\nolimits _ { x \in X_2} f(x)$. According to Lemma 5 of~\cite{xu2018how}, when $X_1 \neq X_2$, $\sum\nolimits _ { x \in X_1} f(x) \neq \sum\nolimits _ { x \in X_2} f(x)$. So $h_1(c, X_1) \neq h_1(c, X_2)$.

For $h_2$, we have $h_2(c, X_1) - h_2(c, X_2) = (\big\vert X_1 \big\vert - \big\vert X_2 \big\vert) \cdot H$. As $\alpha_{cx}>0$ due to the softmax function, and $f(x)>0$ in our example, we know $H>0$. Moreover as $\big\vert X_1 \big\vert - \big\vert X_2 \big\vert \neq 0$, we can get $h_2(c, X_1) \neq h_2(c, X_2)$.
\end{proof}

\section{Details of Datasets}

For the node classification task, we generate a graph with 4800 nodes and 32400 edges. $40.58 \%$ of the nodes are included in triangles as vertices while $59.42 \%$ are not. There are 4000 nodes assigned with feature '0', 400 with feature '1' and 400 with feature '2'. The label of each node for prediction is whether or not it's included in a triangle.

For the graph classification task, detailed statistics of the bioinformatics and social network datasets are listed in Table \ref{Detail}. All of the datasets are available at https://ls11-www.cs.tu-dortmund.de/staff/morris/graphkerneldatasets.

In all datasets, if the original node features are provided, we use the one-hot encodings of them as input.

\begin{table}[H]
\centering
\scriptsize
\caption{Dataset Description} \label{Detail}
\smallskip
\begin{tabular}{lccccc}
\toprule
Datasets  &{Graphs}&{Classes}&{Features}&{Node Avg.}&{Edge Avg.}\\
\midrule
{MUTAG}    & 188   & 2  & 7  & 17.93  & 19.79 \\
{PROTEINS} & 1113  & 2  & 4  & 39.06  & 72.81\\
{ENZYMES}  & 600   & 6  & 6  & 32.63  & 62.14 \\
{NCI1}     & 4110  & 2  & 23 & 29.87  & 32.30 \\
{RE-B}     & 2000  & 2  & -  & 429.63 & 995.51 \\
{RE-M5K}   & 4999  & 5  & -  & 508.52 & 1189.75 \\
\bottomrule
\end{tabular}
\end{table}

\section{Details of Experiment Settings}
For all experiments, we perform 10-fold cross-validation and repeat the experiments 10 times for each dataset and each model. To get a final accuracy for each run, we select the epoch with the best cross-validation accuracy averaged over all 10 folds. The average accuracies and their standard deviations are reported based on the results across the folds in all runs.

In our \textit{Additive} and \textit{Scaled} models, all MLPs have 2 layers with ReLU activation.

In the GAT variants, we use 2 GNN layers and a hidden dimensionality of 32. The negative input slope of $\operatorname { LeakyReLU }$ in the $\operatorname { GAT }$ attention mechanism is 0.2. The number of heads in multi-head attention is 1.

In the GAT-GC variants, we use 4 GNN layers. For the $\operatorname{ Readout }$ function in all models, we use sum for bioinformatics datasets and mean for social network datasets. We apply Batch normalization~\cite{ioffe2015batch} after every hidden layers. The hidden dimensionality is set as 32 for bioinformatics datasets and 64 for social network datasets. The negative input slope of $\operatorname { LeakyReLU }$ in the $\operatorname { GAT }$ attention mechanism is 0.2. We use a single head in the multi-head attention in all models.

All models are trained using the Adam optimizer~\cite{kingma2014adam} and the learning rate is dropped by a factor of 0.5 every 400 epochs in the node classification task and every 50 epochs in the graph classification task. We use an initial learning rate of 0.01 for the TRIANGLE-NODE and bioinformatics datasets and 0.0025 for the social network datasets. For the GAT variants, we use a dropout ratio of 0 and a weight decay value of 0. For the GAT-GC variants on each dataset, the following hyper-parameters are tuned: (1) Batch size in $\{32, 128\}$; (2) Dropout ratio in $\{0, 0.5\}$ after dense layer; (3) $L_2$ regularization from $0$ to $0.001$. On each dataset, we use the same hyper-parameter configurations in all model variants for a fair comparison.

For the results of the baselines for comparison, we use the results reported in their original works by default. If results are not available, we use the best testing results reported in~\cite{xinyi2019capsule,ivanov2018anonymous}.

\end{document}